\documentclass[twoside]{article}

\usepackage[accepted]{aistats2025}
\usepackage[round]{natbib}

\usepackage{hyperref}       % hyperlinks
\usepackage{booktabs}       % professional-quality tables
\usepackage{amsfonts}       % blackboard math symbols
\usepackage{nicefrac}       % compact symbols for 1/2, etc.
\usepackage{microtype}      % microtypography
\usepackage{amsmath}
\usepackage{graphicx}
\usepackage{caption}
\usepackage{subcaption}
\usepackage[T1]{fontenc}

\begin{document}

\graphicspath{ {figures/} }

% If your paper is accepted and the title of your paper is very long,
% the style will print as headings an error message. Use the following
% command to supply a shorter title of your paper so that it can be
% used as headings.
%

\runningtitle{Efficient 3D Molecular Generation}

% If your paper is accepted and the number of authors is large, the
% style will print as headings an error message. Use the following
% command to supply a shorter version of the authors names so that
% they can be used as headings (for example, use only the surnames)
%
%\runningauthor{Surname 1, Surname 2, Surname 3, ...., Surname n}

\runningauthor{Ross Irwin, Alessandro Tibo, Jon Paul Janet, Simon Olsson}

\twocolumn[

\aistatstitle{\textsc{SemlaFlow} -- Efficient 3D Molecular Generation with \\Latent Attention and Equivariant Flow Matching}

\aistatsauthor{ Ross Irwin$^{\textnormal{1,2}}$\footnotemark[1]  \And  Alessandro Tibo$^{\textnormal{1}}$ \And Jon Paul Janet$^{\textnormal{1}}$ \And Simon Olsson$^\textnormal{2}$ }

\vspace{10pt}

\aistatsaddress{
$^{1}$Molecular AI \\
Discovery Sciences, R\&D \\
AstraZeneca \\
Gothenburg, Sweden
\And
$^{2}$Department of Computer Science and Engineering \\
Chalmers University of Technology \\
and University of Gothenburg \\
Gothenburg, Sweden
}]

\begin{abstract}
Methods for jointly generating molecular graphs along with their 3D conformations have gained prominence recently due to their potential impact on structure-based drug design. Current approaches, however, often suffer from very slow sampling times or generate molecules with poor chemical validity. Addressing these limitations, we propose \textsc{Semla}, a scalable E(3)-equivariant message passing architecture. We further introduce an unconditional 3D molecular generation model, \textsc{SemlaFlow}, which is trained using equivariant flow matching to generate a joint distribution over atom types, coordinates, bond types and formal charges. Our model produces state-of-the-art results on benchmark datasets with as few as 20 sampling steps, corresponding to a two order-of-magnitude speedup compared to state-of-the-art. Furthermore, we highlight limitations of current evaluation methods for 3D generation and propose new benchmark metrics for unconditional molecular generators. Finally, using these new metrics, we compare our model's ability to generate high quality samples against current approaches and further demonstrate \textsc{SemlaFlow}'s strong performance.
\end{abstract}

\section{INTRODUCTION}

Generative models for 3D drug design have recently seen a surge of interest due to their potential to design binders for given protein pockets. Some recently proposed models have attempted to directly generate ligands within rigid pockets~\citep{sbdd:pocket2mol,sbdd:target-diff,sbdd:diff-sbdd}. More thorough analysis, however, revealed that many of these models generate ligands with unrealistic binding poses~\citep{sbdd:posebusters,sbdd:benchmarking}. Others have attempted to train unconditional 3D molecular generators as a starting point~\citep{model:edm,model:equi-fm,model:midi,model:mudiff,model:gcdm,model:gfm-diff,model:eqgatdiff}. Specifically, models which apply diffusion~\citep{diff:ddpm,diff:score-matching} to molecular coordinates have been particular popular. However, these models also suffer significant practical limitations; namely, they almost all require hundreds or even thousands of forward passes during generation, making them impractical for most downstream applications. Many also generate chemically unrealistic or poor quality samples when applied to datasets of drug-like molecules.

\footnotetext[1]{Correspondence to \texttt{rossir@chalmers.se}}

\begin{figure*}[t]
    \centering
    \includegraphics[width=\linewidth]{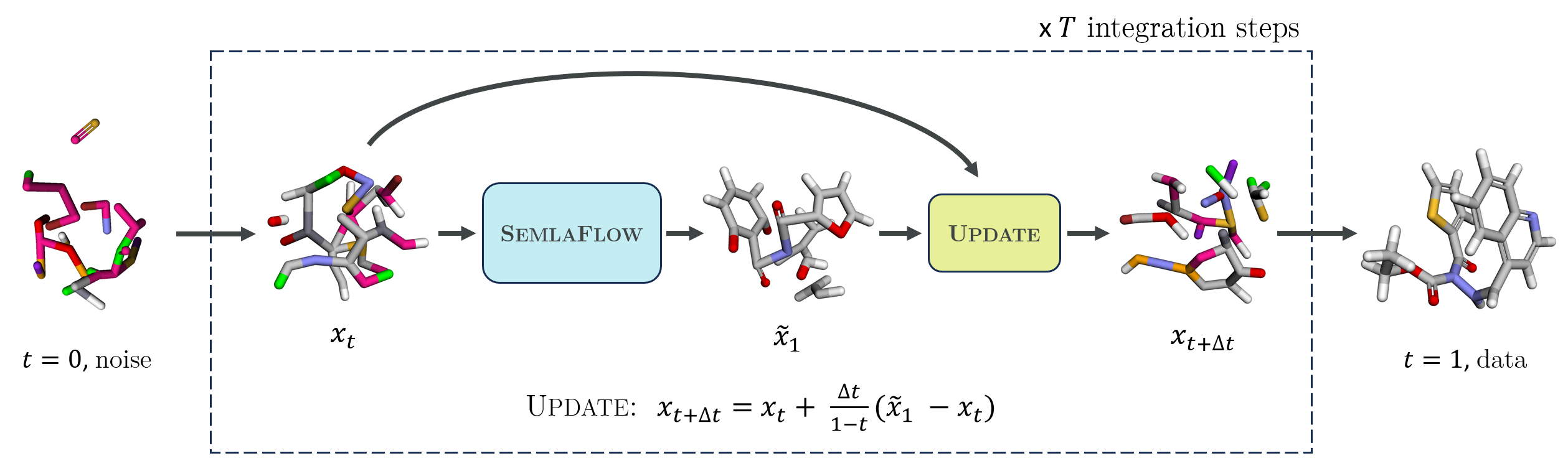}
    \caption{An overview of the inference procedure for \textsc{SemlaFlow}. Noise is firstly sampled from the prior distribution over atom types, bond types and coordinates. We then integrate the vector field by applying our current prediction to the \textsc{SemlaFlow} model and taking a step in the direction of the model's prediction. To improve readability noisy samples are displayed with bonds inferred based on the 3D coordinates.}
    \label{fig:abstract}
\end{figure*}

For molecular generators which represent molecules as strings or 2D graphs, fine-tuning for specific protein pockets has proven very fruitful~\citep{model:reinvent2,model:reinvent4,Atance_2022} and is currently standard practice in the field. Frequently, these models are guided into optimised chemical spaces using reinforcement learning (RL). This approach, while very effective, requires that high quality molecules can be sampled very quickly. Existing 3D molecular generators, which use fully-connected message passing, exhibit poor scaling to larger molecules and larger model sizes -- state-of-the-art unconditional generators~\citep{model:midi,model:eqgatdiff} take minutes to sample a single batch, making them impractical for RL-based fine-tuning.

In this work we tackle this problem from two directions. Firstly, we introduce a novel equivariant architecture for 3D graph generation, \textsc{Semla}, where attention between nodes is applied in a reduced latent space. Our architecture exhibits significantly better efficiency and scalability than existing approaches; even with twice as many parameters as the current state-of-the-art, our model processes a batch of molecules more than three times as quickly. Secondly, we propose a flow matching model for 3D molecular generation which learns a joint distribution over the molecular topology, atomic coordinates and formal charges. Our model, \textsc{SemlaFlow}, provides state-of-the-art results with as few as 20 sampling steps, corresponding to a 2-order-of-magnitude increase in sampling speed compared to existing models. Finally, we also highlight issues with frequently used evaluation metrics for unconditional 3D generation and propose the use of energy and strain energy for benchmarking unconditional 3D generative models.

We provide an overview of our molecular generation approach in Figure~\ref{fig:abstract} and we summarise our key contributions as follows:
\begin{itemize}
    \item \textsc{\textbf{Semla}} -- We introduce a novel E(3)-equivariant architecture, \textsc{Semla}, with significantly better scalability and efficiency than previous molecular generation approaches.
    \item \textsc{\textbf{SemlaFlow}} -- A state-of-the-art molecular generator trained using flow matching with equivariant optimal transport, which provides a more than 100-fold improvement in sampling time compared to existing approaches while maintaining state-of-the-art performance.
    \item \textbf{Benchmarking} -- We introduce new evaluation metrics to address a number of shortcomings with existing metrics for 3D molecular generation.
\end{itemize}

\section{BACKGROUND}

\paragraph{Flow Matching}

Flow matching seeks to learn a generative process which transports samples from a noise distribution \(p_{\mathrm{noise}}\) to samples from a data distribution \(p_{\mathrm{data}}\). Conditional flow matching (CFM) has emerged in different flavors as an effective way to train flow matching models~\citep{fm:stochastic-interpolants,fm:rectified-flow,fm:gaussian-fm}. CFM works in a simulation-free manner by interpolating between noise and a data sample \(x_1 \sim p_\mathrm{data} (\mathbf{x}_1)\). To do this a time-dependent conditional flow \(p_{t|1}(\cdot|\mathbf{x}_1)\) is defined, from which a conditional vector field \(u_t(\mathbf{x}_t|\mathbf{x}_1)\) can be derived. Typically, a model \(v_t^\theta (\mathbf{x}_t)\) is trained to regress the vector field, but other formulations~\citep{fm:discrete-fm,model:flowsite} have trained a model to estimate the distribution \(p_{1|t}^\theta(\cdot|\mathbf{x}_t)\), which reconstructs clean data from noisy data. The vector field can then be constructed using the expectation:
\begin{equation}
v_t^\theta (\mathbf{x}_t) = \mathbb{E}_{\mathbf{\tilde x}_1 \sim p_{1|t}^\theta (\mathbf{x}_1 | \mathbf{x}_t)} (\left[ u_t (\mathbf{x}_t | \mathbf{\tilde x}_1) \right])
\end{equation}
Samples can then be generated by integrating the vector field with an arbitrary ODE solver.

\paragraph{Invariance and Equivariance}

Group invariance and equivariance are crucial properties to consider when designing models for 3D molecular generation. For a group \(\mathcal{G}\), if \(T_g\) and \(P_g\) are linear representations of a group element \(g \in \mathcal{G}\), then a probability density \(p(\mathbf{x})\) is considered \textit{invariant} with respect to \(\mathcal{G}\) iff \( p(T_g \mathbf{x}) = p(\mathbf{x}) \) for all \(g \in \mathcal{G}\), and a function \(f\) is considered \textit{equivariant} to \(\mathcal{G}\) iff \( T_g(f(\mathbf{x})) = f(P_g(\mathbf{x})) \) for all \(g \in \mathcal{G}\).

\citet{fm:equi-flows} showed that, if a base density \(p_0(\mathbf{x}_0)\) is \(\mathcal{G}\)-invariant and a target density \(p_1(\mathbf{x}_1)\) is generated by following a \(\mathcal{G}\)-equivariant vector field, then \(p_1(\mathbf{x}_1)\) is also \(\mathcal{G}\)-invariant. We use this finding to ensure that the density of molecular coordinates learned by our model is \(\mathcal{G}\)-invariant by only applying equivariant updates and sampling coordinate noise from an isotropic Gaussian. For molecular generation we are concerned with the group \(\mathcal{G} = \mathrm{E}(3) \times S_N\) where \(\mathrm{E}(3)\) is the Euclidean group in 3 dimensions, encompassing translations, rotations and reflections, and \(S_N\) is the symmetric group for a set with \(N\) elements -- the group of all possible permutations.

\section{The \textsc{Semla} Architecture}
\label{section:semla}

Existing state-of-the-art models for 3D molecular generation~\citep{model:edm,model:midi,model:eqgatdiff} use fully-connected, multi-layer perceptron (MLP)-based message passing layers. However, the computational cost of such layers scales quadratically in both the feature dimension and the number of atoms. Consequently, these layers become a significant computational bottleneck when scaling to larger, drug-like molecules.

\begin{figure}
    \centering
    \includegraphics[width=0.85\linewidth]{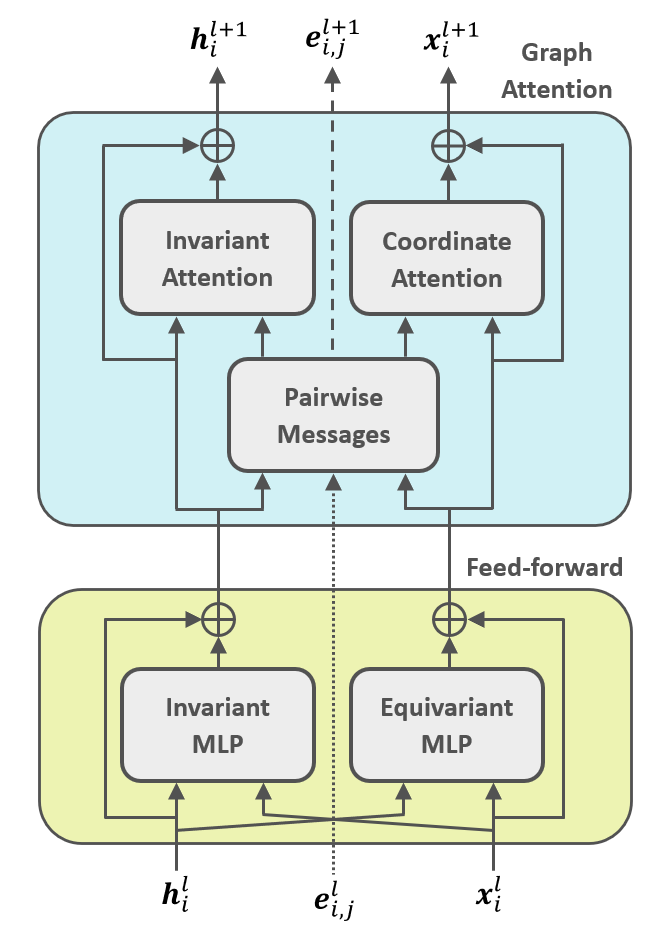}
     \caption{Architectural overview of one \textsc{Semla} layer.}
     \label{fig:layer-overview}
\end{figure}

To alleviate this problem we propose \textsc{Semla} - a {\underline s}calable {\underline e}quivariant model which uses {\underline m}ulti-head {\underline l}atent graph {\underline a}ttention, where message passing is performed on compressed latent representations. This extension allows us to scale the dimensionality of the node features and the number of learnable model parameters without leading to prohibitory increases in computational cost. We illustrate one \textsc{Semla} architecture in Figure~\ref{fig:layer-overview} and further expand on each component below.

Similarly to previous approaches, \textsc{Semla} makes use of both \(\mathrm{E}(3)\) invariant and equivariant features. Enforcing group symmetry provides strong inductive biases and improves sample complexity~\citep{exploiting-symmetries,inv:inv-densities,model:edm}. However, unlike many existing molecular generation architectures, \textsc{Semla} does not distinguish between molecular coordinates and equivariant feature vectors, but rather treats them as a single learnable representation. We argue that this representation has two key benefits over previous molecular generation approaches. Firstly, sets of learnable equivariant features provide much more expressivity than models which store only one set of coordinates per molecule~\citep{architecture:egnn,model:edm,model:midi}. Additionally, a joint representation of equivariant features allows for a simpler update mechanism than that proposed in EQGAT~\citep{architecture:eqgat} -- we can simply apply linear projections (without bias) to create and update the feature vectors while maintaining equivariance.

To ensure stable training we use normalisation layers throughout the model. LayerNorm~\citep{dl:layer-norm} is applied to invariant features, and, for equivariant features, we adapt the normalisation scheme from MiDi~\citep{model:midi} to allow for multiple equivariant feature vectors. We hypothesise that this allows the model to learn equivariant features of different length scales, which helps to circumvent the problem of molecules of different sizes being normalised to have the same average vector norm. We further extend the normalisation to ensure that coordinates are zero-centred. An additional zero-centering, which we apply at the end of the model, ensures that the learned density is translation invariant~\citep{model:equi-normflows,model:geodiff}. For the remainder of this paper we will use \(\phi_{inv}(\cdot)\) and \(\phi_{equi}(\cdot)\) to refer to the normalisation functions for invariant and equivariant features, respectively. Both LayerNorm and our coordinate normalisation module contain a small number of learnable parameters, these are not shared between layers.

We will denote invariant and equivariant features for an atom \(i\) as \(\mathbf{h}_i \in \mathbb{R}^{d_{inv}}\) and \(\mathbf{x}_i \in \mathbb{R}^{d_{equi} \times 3}\), respectively, where \(d_{inv}\) is the number of invariant scalar features and \(d_{equi}\) is the number of equivariant feature vectors. We also use \(N\) to refer to the number of atoms in the molecule. To simplify the notation, we assume that operations applied to \(\mathbf{x}_i\) implicitly correspond to the concatenation of the results of the operation applied to individual vectors, unless we make the equivariant feature vector explicit using a superscript. For example, the norm of \(\mathbf{x}_i\) is implicitly applied as \( \lVert \mathbf{x}_i \rVert = \left[ \lVert \mathbf{x}_i^1 \rVert, \lVert \mathbf{x}_i^2 \rVert, \dots , \lVert \mathbf{x}_i^{d_{equi}} \rVert \right] \).

\subsection{Feature Feed-forward}

The feed-forward component provides a simple feature update mechanism while also allowing the exchange of information between invariant and equivariant features. The feed-forward update is given as follows:
\begin{gather}
    \mathbf{\tilde h}_{k} = \phi_{inv}(\mathbf{h}_{k})
    \hspace{25pt}
    \mathbf{\tilde x}_{k} = \mathbf{W}_\theta^1 \phi_{equi}(\mathbf{x}_{k})
    \\
    \mathbf{h}_{i}^{\text{ff}} = \mathbf{h}_{i} + \Phi_{\theta} (
    \mathbf{\tilde h}_{i} , \lVert \phi_{equi}(\mathbf{x}_{i}) \rVert )
    \\
    \mathbf{x}_{i}^{\text{ff}} = \mathbf{x}_{i} +
    \mathbf{W}_\theta^2 \left(
    \sum_{j=1}^{d_{equi}} \mathbf{\tilde x}_i^j \otimes 
    \Psi_{\theta}(\mathbf{\tilde h}_{i}) \right)
\end{gather}
where \(\Phi_{\theta}\) and \(\Psi_{\theta}\) are learnable multi-layer perceptrons, \(\mathbf{W}_\theta^1 \in \mathbb{R}^{d_{equi} \times d_{equi}}\) and \(\mathbf{W}_\theta^2 \in \mathbb{R}^{d_{equi} \times d_{equi}}\) are learnable weight matrices, and \(\otimes\) is the outer product. Similarly to the transformer architecture~\citep{architecture:transformer}, \(\Phi_\theta\) linearly maps features to \(4 \times d_{inv}\), applies a non-linearity (we use SiLU~\citep{dl:silu} throughout) and then maps back to \(d_{inv}\).

\subsection{Equivariant Graph Attention}

In this section, we introduce a novel attention mechanism for 3D graph structures. Like previously proposed attention mechanisms for equivariant architectures~\citep{architecture:egnn,architecture:eqgat,architecture:equiformer}, our model computes pairwise messages using a multi-layer perceptron (MLP). These pairwise messages are then split in two and passed to attention mechanisms for invariant and equivariant features, respectively. We describe each of these components in more detail below.

\paragraph{Latent Message Passing}

Pairwise messages are computed using a 2-layer MLP which combines invariant node features with pairwise dot products from the equivariant features. Similarly to previous approaches we compute messages between all pairs of nodes in the graph. Unlike models such as EGNN~\citep{architecture:egnn}, MiDi~\citep{model:midi} and EQGAT~\citep{architecture:eqgat}, however, we first compress the invariant node features into a smaller latent space, with dimensionality \(d_{l}\), using a learnable linear map. This reduces the computational complexity of the pairwise MLP from \(\mathcal{O} (N^2 d_{inv}^2)\) to \(\mathcal{O} (N^2 d_{l}^2)\) where \(d_l \ll d_{inv}\), leading to a significant reduction in the compute and memory overhead of the MLP, especially on larger molecules. It also allows us to scale the size of the invariant node features independently of the node latent dimension. In Appendix~\ref{section:ablation-results} we provide experiments varying the size of this parameter and show that it is possible to produce a significant reduction in inference time with negligible drop in generation quality. Formally, messages between nodes \(i\) and \(j\), which are split into invariant and equivariant attention components, are computed as follows:
\begin{gather}
    ( \mathbf{m}_{i,j}^{(inv)} ,  \mathbf{m}_{i,j}^{(equi)} )
    = \Omega_{\theta} (
    \mathbf{\tilde h}_{i}, \mathbf{\tilde h}_{j},
    \mathbf{\tilde x}_{i} \cdot \mathbf{\tilde x}_{j}
    )
    \\
    \mathbf{\tilde h}_k = \mathbf{W}_\theta^3 \phi_{inv}(\mathbf{h}_k^{\text{ff}})
    \hspace{25pt}
    \mathbf{\tilde x}_k = \phi_{equi}(\mathbf{x}_k^{\text{ff}})
\end{gather}
where \(\Omega_\theta\) is the pairwise message MLP and \(\mathbf{W}_\theta^3 \in \mathbb{R}^{d_l \times d_{inv}}\) is a learnable latent projection matrix.

\paragraph{Invariant Feature Attention}

Once messages have been computed a softmax operation is applied to produce attention weights between pairs of nodes. These weights are then used to aggregate node features by taking a weighted average. Since the message vectors can, in general, be smaller than the node features, each scalar in the message vector attends to a fixed number of scalars within the node feature vectors. We note that this attention implementation generalises the attention mechanism found in EQGAT and related models such as the Point Transformer~\citep{architecture:point-transformer}, where each scalar in the message attends to exactly one scalar in the node features, and is very closely related to the multi-head attention mechanism adopted in GAT~\citep{architecture:gat,architecture:gatv2} and the Transformer~\citep{architecture:transformer}. We also make use of the recently proposed variance preserving aggregation mechanism~\citep{dl:gnn-vpa}, which corresponds to multiplying the attended vectors by weights \(w_i^k\). Overall, our invariant feature attention is computed as:
\begin{gather}
    \alpha_{i,j}^k = \frac{\text{exp}(m_{i,j}^{k, (inv)})}{\sum_{j^\prime = 1}^N \text{exp}(m_{i,j^\prime}^{k, (inv)})}
    \hspace{15pt}
    w_i^k = \sqrt{\sum_{j=1}^N (\alpha_{i,j}^k)^2}
    \\
    \mathbf{\tilde h}_i = \mathbf{W}_\theta^4 \phi_{inv}(\mathbf{h}_i^{\text{ff}})
    \hspace{25pt}
    \mathbf{a}_i^k = \sum_{j = 1}^{N} \alpha_{i,j}^k \mathbf{\tilde h}_j^{k}
    \\
    \mathbf{h}_i^{\text{out}} = \mathbf{h}_i^{\text{ff}} +
    \mathbf{W}_\theta^5 \Big( \Big\|_{k=1}^{K} w_i^k \mathbf{a}_i^k \Big)
\end{gather}
where \( \mathbf{W}_\theta^4 \in \mathbb{R}^{d_{inv} \times d_{inv}} \) and \( \mathbf{W}_\theta^5 \in \mathbb{R}^{d_{inv} \times d_{inv}} \) are learnable weight matrices and \(\|\) is the concatenation operation. Here, node features are split into \(n_{heads}\) equally sized segments and each scalar attention score \(\alpha_{i,j}^k\) attends to one segment \(\mathbf{\tilde h}_j^{k}\).

\paragraph{Equivariant Feature Attention}

Similarly to the invariant features, messages for the equivariant features are used to apply an attention-based update. The attention function applied here is line with previous work~\citep{architecture:egnn,model:midi,architecture:eqgat}, however we extend it to allow for multiple equivariant feature vectors. Notably, we also find the use of softmax normalisation on raw messages to be beneficial for overall model performance. Analogously to the invariant feature attention, we also apply variance preserving updates to the equivariant features. \textsc{Semla} attention for equivariant features is therefore defined as follows:
\begin{gather}
    \mathbf{\hat{x}}_{i,j} = \frac{\mathbf{\tilde x}_j - \mathbf{\tilde x}_i}
    {\lVert \mathbf{\tilde x}_j - \mathbf{\tilde x}_i \rVert}
    \hspace{25pt}
    \mathbf{\tilde x}_{k} = \mathbf{W}_\theta^6 \phi_{equi}(\mathbf{x}_{k}^{\text{ff}})
    \\[5pt]
    \alpha_{i,j}^k = \frac{\text{exp}(m_{i,j}^{k, (equi)})}{\sum_{j^\prime = 1}^N \text{exp}(m_{i,j^\prime}^{k, (equi)})}
    \\[5pt]
    a_i^k = \sum_{j=1}^N \alpha_{i,j}^k \hat{x}_{i,j}^k
    \hspace{25pt}
    w_i^k = \sqrt{\sum_{j=1}^N (\alpha_{i,j}^k)^2}
    \\[5pt]
    \mathbf{x}_{k}^{\text{out}} = \mathbf{x}_{k}^{\text{ff}} +
    \mathbf{W}_\theta^7 \Big( \left[ w_i^1 a_i^1 , \dots , w_i^K a_i^K \right]^T \Big)
\end{gather}

\subsection{Overall Architecture}

A full \textsc{Semla} model consists of a stack of \textsc{Semla} layers along with embedding layers and MLPs for encoding the atom and bond types, and MLP prediction heads for producing unnormalised distributions for atoms, bonds and formal charges. Unlike EQGAT, our model does not carry edge features throughout the network. Instead, the first layer embeds bond information into the node features by passing the encoded bond features into the pairwise message module. Analogously, the final layer produces pairwise edge features which are then further updated through a bond refinement layer at the end of the network. This layer acts in a similar way to the pairwise message block described above but only updates the edge features. Absorbing bond information into the node features like this leads to a further increase in the efficiency of our model and, in our experiments, had little impact on generative performance.

Unless stated otherwise, the \textsc{Semla} models we present in the remainder of this paper are constructed from 12 layers with \(d_{inv} = 384\), \(d_{equi} = 64\), \(d_{l} = 64\), \(n_{heads} = 32\). This corresponds to approximately 22M learnable parameters. We provide a full model overview and further hyperparameter details in Appendix~\ref{section:training-appendix}.

\section{FLOW MATCHING FOR MOLECULAR GENERATION}

To assess the ability of \textsc{Semla} to model distributions with $E(3)\times S_N$ symmetry we apply conditional flow matching~\citep{fm:gaussian-fm,fm:stochastic-interpolants,fm:stochastic-interpolants-full,fm:rectified-flow} with optimal transport to create a generative model for molecules, which we refer to as \textsc{SemlaFlow}. Unlike many previous approaches to 3D generation, we learn to generate a join distribution over atomic coordinates, atom types, bond orders and formal charges, rather than inferring parts of this distribution after generation.

Each molecule is represented by a tuple \(z = ( \mathbf{x}, \mathbf{a}, \mathbf{b}, \mathbf{c} ) \) of coordinates, atom types, bond orders and formal charges, respectively. Since we wish to train a model to sample from the joint distribution \(p (z)\) which contains both discrete and continuous data, we parameterise a single \textsc{Semla} model to generate multiple vector fields. For discrete data (atom and bond types) we apply the discrete flow models (DFM) framework~\citep{fm:discrete-fm}, and for the continuous atom coordinates we apply the flow matching algorithm proposed by~\citet{ot:minibatch-ot}, where a small amount of gaussian noise is added the interpolated coordinates \(x_t\). Our model also predicts the formal charge for each atom but these do not participate in the generative flow matching process. In the remainder of this section we outline the full training and sampling procedure for \textsc{SemlaFlow} with molecular structures.

\paragraph{Training \textsc{SemlaFlow}}

As shown in existing conditional flow matching frameworks~\citep{fm:gaussian-fm,fm:stochastic-interpolants,fm:discrete-fm}, training proceeds by firstly sampling: noise \(z_0 \sim p_{\mathrm{noise}} (z_0) \); data \(z_1 \sim p_{\mathrm{data}} (z_1) \); and a time \( t \in \left[ 0, 1 \right] \), and using these to sample from the time-dependent conditional flow \( z_t \sim p_{t|1} (z | z_0, z_1) \). The joint molecular interpolation is therefore given as follows:
\begin{gather}
    \mathbf{x}_t \sim \mathcal{N} ( t \mathbf{x}_1 + (1-t) \mathbf{x_0} , \sigma^2 )
    \hspace{15pt}
    t \sim \text{Beta}(\alpha, \beta)
    \\
    a_t \sim \text{Cat} ( t \delta \{ a_1, a_t \} + (1-t) \frac{1}{|\mathcal{A}|} )
    \\
    b_t \sim \text{Cat} ( t \delta \{ b_1, b_t \} + (1-t) \frac{1}{|\mathcal{B}|} )
\end{gather}
Where \(\mathcal{A}\) and \(\mathcal{B}\) are the sets of atom and bond types, respectively, and \(\delta \{ i, j \} \) is the Kronecker delta which is $1$ when $i = j$ and $0$ otherwise. We use \((\alpha, \beta) = (2.0, 1.0)\) and \(\sigma = 0.2\) for all \textsc{SemlaFlow} models presented in this paper. In practice, we apply the equivariant optimal transport (OT)~\citep{ot:equi-ot,model:equi-fm} transformation to the sampled coordinates \( \hat{\mathbf{x}}_0 = f_{\pi} (\mathbf{x}_0, \mathbf{x}_1) \) before sampling the interpolated value \(\mathbf{x}_t\). This corresponds to applying a permutation and rotation which minimises the transport cost (in this case the mean-squared error) between \(\mathbf{x}_0\) and \(\mathbf{x}_1\).

%%% This will have to stay here to ensure it goes in right place

\begin{table*}[ht!]
  \caption{Molecular generation results on QM9. Models are grouped into those which infer bonds from coordinates (top) and those which generate bonds directly (bottom). Since some models only publish the proportion of molecules which are both unique and valid, results marked $^*$ are estimates for uniqueness.}
  \label{table:qm9-results}
  \centering
  \begin{tabular}{lccccc}
    \toprule
    Model     & Atom Stab $\uparrow$  & Mol Stab $\uparrow$  & Valid $\uparrow$ & Unique $\uparrow$  & NFE  \\
    \midrule
    EDM       & 98.7  & 82.0  & 91.9  & 98.9$^*$  & 1000   \\
    GCDM      & 98.7  & 85.7  & 94.8  & 98.4$^*$  & 1000   \\
    GFMDiff   & 98.9  & 87.7  & 96.3  & 98.8$^*$  & 500    \\
    EquiFM    & 98.9  & 88.3  & 94.7  & 98.7$^*$  & 210    \\
    GeoLDM    & 98.9  & 89.4  & 93.8  & 98.8      & 1000   \\
    MUDiff    & 98.8  & 89.9  & 95.3  & 99.1      & 1000   \\
    GeoBFN    & 99.3  & 93.3  & 96.9  & 95.4      & 2000   \\
    \midrule
    FlowMol   & 99.7  & 96.2  & 97.3  & --        & 100    \\
    MiDi      & 99.8  & 97.5  & 97.9  & 97.6      & 500    \\
    EQGAT-diff       & \textbf{99.9}$_{\pm \text{0.0}}$  & 98.7$_{\pm \text{0.18}}$
        & 99.0$_{\pm \text{0.16}}$ & \textbf{100.0}$_{\pm \text{0.0}}$  & 500        \\
    \textsc{SemlaFlow} (Ours)  & \textbf{99.9}$_{\pm \text{0.0}}$  & \textbf{99.7}$_{\pm \text{0.03}}$
        & \textbf{99.4}$_{\pm \text{0.03}}$ & 95.4$_{\pm \text{0.12}}$  & 100        \\
    \bottomrule
  \end{tabular}
\end{table*}

Previous work on molecular structure generation has found it beneficial to train models to predict data directly rather than noise~\citep{model:eqgatdiff} or a vector field~\citep{model:flowsite}. \textsc{SemlaFlow} is therefore trained to learn a distribution \( p_{1|t}^\theta (z_1 | z_t ) \) which predicts clean data from interpolated data using a \textsc{Semla} model with parameters \(\theta\). After sampling a predicted molecule \( {\tilde z}_1 \sim p_{1|t}^\theta (z_1 | z_t ) \) where \({\tilde z}_1 = (\mathbf{\tilde x}_1, \mathbf{\tilde a}_1, \mathbf{\tilde b}_1, \mathbf{\tilde c}_1) \), the model is trained with a mean-squared error loss function (\(\mathcal{L}_{\mathrm{MSE}}\)) for coordinates, and cross-entropy losses (\(\mathcal{L}_{\mathrm{CE}}\)) for atom types, bond types and charges. The final loss for the model is then given by the weighted sum:
\begin{gather}
    \mathcal{L}_{\mathrm{SemlaFlow}} = \lambda_x \mathcal{L}_{\mathrm{MSE}}(\mathbf{\tilde x}_1 , \mathbf{x}_1)
    + \lambda_a \mathcal{L}_{\mathrm{CE}}(\mathbf{\tilde a}_1 , \mathbf{a}_1)
    \\
    + \lambda_b \mathcal{L}_{\mathrm{CE}}(\mathbf{\tilde b}_1 , \mathbf{b}_1)
    + \lambda_c \mathcal{L}_{\mathrm{CE}}(\mathbf{\tilde c}_1 , \mathbf{c}_1)
\end{gather}

We also make use of self-conditioning, which was originally proposed for diffusion models as way of reusing the model's previous prediction when sampling~\citep{diff:self-cond}. To create a self-conditioned \textsc{SemlaFlow} model, we adopt the same training procedure as HarmonicFlow~\citep{model:flowsite}. We provide further details on this, along with the hyperparameters used for \textsc{SemlaFlow}, in Appendix~\ref{section:training-appendix}.

\paragraph{Sampling Molecules}

Once we have trained a \textsc{SemlaFlow} model, molecules can be generated by, firstly, sampling noise \(z_0 \sim p_{\mathrm{noise}} (z_0 | n) \), and then integrating the ODE corresponding to the conditional flow \(p_{t|1}\) from \(t = 0\) to \(t = 1\). For coordinates, the vector field corresponding to our choice of conditional flow is given by \( \mathbf{\tilde x}_1 - \mathbf{x}_0 = \frac{1}{1 - t} ( \mathbf{\tilde x}_1 - \mathbf{x}_t ) \) where \(\mathbf{\tilde x}_1 \sim p_{1|t}^\theta\) as shown above. We then apply an Euler solver to integrate the ODE with step sizes \(\Delta t\) as follows: \( \mathbf{x}_{t + \Delta t} = \mathbf{x}_t + \frac{\Delta t}{1 - t} ( \mathbf{\tilde x}_1 - \mathbf{x}_t ) \). We refer readers to DFM~\citep{fm:discrete-fm} for the sampling procedure for atom and bond types. In practice, we found that taking logarithmically spaced steps, where the model spends more time in parts of the vector field closer to \(t = 1\), resulted in better performance than using constant step sizes.

\section{EXPERIMENTS}

In this section we provide results on benchmark 3D molecular generation tasks and compare the performance of our model to existing state-of-the-art approaches. In Appendix~\ref{section:ablation-results} we also provide results on multiple ablation experiments which demonstrate the impact of various contributions of this paper. One experiment demonstrates how controlling the latent message size \(d_l\) can lead to a significant model speedup without impacting performance. Another provides a side-by-side comparison of the \textsc{Semla} architecture with the EQGAT and EGNN architectures trained using identical flow matching training setups. We show that \textsc{Semla} outperforms the EQGAT and EGNN architectures at various model sizes, while also providing between 3 and 5 times faster inference over larger models. Finally, we provide samples from a \textsc{SemlaFlow} model trained on GEOM Drugs in Appendix~\ref{section:samples}.

\paragraph{Evaluation Setup}

Two benchmark datasets, QM9~\citep{dataset:qm9} and GEOM Drugs~\citep{dataset:geom}, are used to assess \textsc{SemlaFlow}'s abilities as an unconditional molecular generator. Since QM9 contains only very small molecules GEOM Drugs serves a more useful benchmark for distinguishing model performance. For both datasets we use the same data splits as MiDi and EQGAT-diff. To improve training times, however, we discard molecules with more than 72 atoms from the GEOM Drugs training set. This corresponds to about 1\% of the training data; validation and test sets are left unchanged. All metrics for \textsc{SemlaFlow} presented below are calculated by sampling from the distribution of molecule sizes in the test set, and then generating molecules with the sampled number of atoms by integrating the trained ODE.

\begin{table*}[ht!]
  \caption{Molecular generation results on GEOM Drugs. Results for models which perform bond inference are provided in the appendix since they often do not provide results for all standard benchmark metrics.}
  \label{table:geom-results}
  \centering
  \begin{tabular}{lcccccc}
    \toprule
    Model            & Atom Stab $\uparrow$  & Mol Stab $\uparrow$ & Valid $\uparrow$  
        & Unique $\uparrow$  & Novel $\uparrow$  & NFE     \\
    \midrule
    FlowMol          & 99.0           & 67.5  & 51.2  & --               & --             & 100      \\
    MiDi             & \textbf{99.8}  & 91.6  & 77.8  & \textbf{100.0}  & \textbf{100.0}  & 500      \\
    EQGAT-diff       & \textbf{99.8}$_{\pm \text{0.0}}$  & 93.4$_{\pm \text{0.21}}$  & \textbf{94.6}$_{\pm \text{0.24}}$
        & \textbf{100.0}$_{\pm \text{0.0}}$  & 99.9$_{\pm \text{0.07}}$  & 500      \\
    \textsc{SemlaFlow} (Ours)   & \textbf{99.8}$_{\pm \text{0.0}}$  & \textbf{97.3}$_{\pm \text{0.08}}$  & 93.9$_{\pm \text{0.19}}$
        & \textbf{100.0}$_{\pm \text{0.0}}$  & 99.6$_{\pm \text{0.03}}$  & 100      \\
    \bottomrule
  \end{tabular}
\end{table*}

We compare \textsc{SemlaFlow} to a number of recently proposed models for 3D molecular generation. EDM~\citep{model:edm}, GCDM~\citep{model:gcdm}, GFMDiff~\citep{model:gfm-diff}, MUDiff~\citep{model:mudiff}, GeoLDM~\citep{model:geoldm} and GeoBFN~\citep{model:geobfn} are all diffusion-based models which infer bonds from atom positions. We also compare to EquiFM~\citep{model:equi-fm} which uses flow-matching along with equivariant optimal transport to generate atom types and coordinates; bonds are then inferred based on these. FlowMol~\citep{model:flowmol} is a recently proposed flow matching model which learns a joint distribution over atoms and bonds and is probably the most similar existing work to ours. Finally, we also compare to MiDi~\citep{model:midi} and EQGAT-diff~\citep{model:eqgatdiff} which we regard as the existing state-of-the-art. We use standard benchmark evaluation metrics: \textit{atom stability}; \textit{molecule stability}; \textit{validity}; \textit{uniqueness}; and \textit{novelty}, which have been thoroughly described in previous works. We also provide a full description of these metrics in Appendix~\ref{section:eval-issues}. Results for models we evaluated are given as an average over 3 runs, sampling 10,000 molecules on each run, with standard deviations provided in subscripts. We also provide the number of function evaluations (NFE) required to sample one batch of molecules.

\paragraph{Molecular Generation Results}

Table~\ref{table:qm9-results} compares the performance of \textsc{SemlaFlow} with existing approaches on the QM9 dataset. Following~\citet{issues-with-qm9,model:edm} we do not provide novelty scores on QM9. \textsc{SemlaFlow} is trained for 300 epochs on QM9 on a single Nvidia A100 GPU. From the table we can see that \textsc{SemlaFlow} matches or exceeds all models on all metrics other than uniqueness, despite using 5 times fewer sampling steps than MiDi and EQGAT-diff. Our model also outperforms EquiFM and FlowMol, the only other flow matching models in the table.

Table~\ref{table:geom-results} compares \textsc{SemlaFlow}'s performance on GEOM Drugs to existing models which are also trained to generate molecular bonds. \textsc{SemlaFlow} matches or exceeds the performance of existing models on atom and molecule stability and produces only slightly fewer valid molecules than EQGAT-diff. The performance difference between FlowMol and \textsc{SemlaFlow} is also much more noticeable than on QM9; only two-thirds of molecules produced by FlowMol satisfy the molecular stability metric, compared to more than 97\% for \textsc{SemlaFlow}. In addition to requiring significantly fewer evaluation steps than MiDi and EQGAT-diff, our model also requires much less compute for training. \textsc{SemlaFlow} trains for 200 epochs on a single Nvidia A100 GPU, compared to 800 epochs with 4 GPUs for EQGAT-diff.

Due to space constraints and since these models often do not provide results for all standard metrics, we provide results for models which infer bonds in Appendix~\ref{section:additional-results}. Notably, while some of these models have higher validity than \textsc{SemlaFlow}, most have molecular stabilities close to 0 -- they struggle to generate molecules with the correct number of bonds for each atom, especially on larger, drug-like molecules, like those in GEOM Drugs. Since the validity metric only measures whether a molecule can be loaded into RDKit~\citep{rdkit}, we consider molecule stability a more comprehensive metric for this task. We discuss this issue further, along with other problems with existing evaluation metrics, in Appendix~\ref{section:eval-issues}.

\paragraph{Further Evaluation}

\begin{table*}
  \caption{Comparison between EQGAT-diff and \textsc{SemlaFlow} with different numbers of sampling steps. Energy and strain energy are given as an average per atom and are measured in $\mathrm{kcal}\cdot \mathrm{mol}^{-1}$. Sample time is measured by the average number of seconds to generate 1000 molecules. All metrics are averaged over 3 runs.}
  \label{table:head-to-head-results}
  \centering
  \begin{tabular}{lcccccc}
    \toprule
    Model              & Mol Stab $\uparrow$  & Valid $\uparrow$     & Energy $\downarrow$  
        & Strain $\downarrow$  & Sample Time $\downarrow$  & NFE   \\
    \midrule
    EQGAT-diff         & 93.4$_{\pm \text{0.21}}$  & 94.6$_{\pm \text{0.24}}$
        & 3.38$_{\pm \text{0.020}}$  & 3.23$_{\pm \text{0.020}}$  & 2293.0  & 500  \\
    \textsc{SemlaFlow}$_{20}$     & 95.3$_{\pm \text{0.14}}$  & 93.0$_{\pm \text{0.10}}$
        & 2.58$_{\pm \text{0.018}}$  & 1.76$_{\pm \text{0.007}}$  & 20.3    & 20   \\
    \textsc{SemlaFlow}$_{50}$     & 97.0$_{\pm \text{0.21}}$  & 93.9$_{\pm \text{0.12}}$  
        & 2.33$_{\pm \text{0.044}}$  & 1.46$_{\pm \text{0.043}}$  & 49.8    & 50   \\
    \textsc{SemlaFlow}$_{100}$    & 97.3$_{\pm \text{0.08}}$  & 93.9$_{\pm \text{0.19}}$
        & 2.24$_{\pm \text{0.028}}$  & 1.37$_{\pm \text{0.022}}$  & 99.3    & 100  \\
    \midrule
    Data              & 100.0  & 100.0  & 1.12  & 0.36  & --  & -- \\
    \bottomrule
  \end{tabular}
\end{table*}

The validity and stability metrics presented here, however, only measure the topological structure of the molecule (i.e. the molecular graph); they provide no information on the quality of the generated conformations. We can also see that some existing models have already saturated these metrics on both QM9 and GEOM Drugs, warranting the introduction of additional evaluation methods. To further compare model performance and to allow evaluation of 3D generation, we introduce energy per atom and strain energy per atom as new benchmark metrics for this task. The energy measures the quality of a conformer, considering typical bonded and non-bonded interactions. The energy $U(\mathbf{x})$ of a conformation is inversely related to its probability according to the Boltzmann distribution $p(\mathbf{x})=Z^{-1}\exp\left(-U(\mathbf{x})/kT\right)$ where \(T\) is the temperature and \(k\) is the Boltzmann constant. The strain is given by the difference \( U(\mathbf{x}) - U(\mathbf{\tilde x})  \) where \(\mathbf{\tilde x}\) is the {\it relaxed} (i.e. minimised) conformation for \(\mathbf{x}\). Since molecular energy is generally calculated as a sum of atomic energies we normalise both metrics by the number of atoms in each generated molecule. We argue that these metrics provide a very useful overview of the quality of the generated conformations and directly include measurements such as bond lengths and bond angles which have been proposed previously~\citep{model:midi,sbdd:posebusters}. We use RDKit~\citep{rdkit} with an MMFF94~\citep{mmff94} forcefield to calculate the energies and perform the minimisation.

In Table~\ref{table:head-to-head-results} we provide a performance and sampling time comparison between EQGAT-diff and \textsc{SemlaFlow} with varying numbers of ODE integration steps. \textsc{SemlaFlow} produces higher molecule stabilities and better energies and strain energies than EQGAT-diff, even with as few as 20 integration steps, although the validity of molecules produced by EQGAT-diff is slightly higher. Using \textsc{SemlaFlow} with 100 sampling steps corresponds to more than a 20x speedup over EQGAT-diff, and using 20 sampling steps corresponds to a two order-of-magnitude increase in sampling speed. Notably, however, molecules generated by EQGAT-diff have lower minimised energies than \textsc{SemlaFlow}, suggesting that their model is better at finding molecular conformations which have lower energy minima, while our model is better at producing lower strain energies.

\section{RELATED WORK}

\paragraph{3D Molecular Generation}

In addition to the unconditional molecular generators we outlined above, a number of works have attempted to directly generate ligands within protein pockets~\citep{sbdd:pocket2mol,sbdd:target-diff,sbdd:diff-sbdd,model:pilot,model:molsnapper}. However, these models also suffer from the issues we outlined previously, including long-sampling times (100s - 1000s of seconds for 100 molecules~\citep{sbdd:diff-sbdd}) and generating invalid chemical structures or molecules with very high strain energies~\citep{sbdd:benchmarking,sbdd:posebusters}. GraphBP~\citep{sbdd:graphbp}, an autoregressive model for protein-conditioned generation, is able to generate ligands faster, but suffers from poor docking scores in comparison to more recent diffusion models.

\paragraph{Flow Matching for 3D Structures}

Outside of small molecule design flow-matching has recently gained traction with generative models for biomolecules. FoldFlow~\citep{protein:foldflow} and FrameFlow~\citep{protein:frameflow} are both recently introduced flow matching models for protein structure generation. Multiflow~\citep{fm:discrete-fm} attempts to jointly generate protein sequence and structure and introduces the discrete flow models (DFM) framework for flow matching generation of discrete data. \citet{fm:dirichletfm} also introduce a framework for flow matching on discrete data, DirichletFM, and apply this to DNA sequence design. Finally, \citet{protein:abode} use a conjoined system of ODEs to train a model to jointly generate antibody sequences and structures.

\section{CONCLUSION}

In this work we have presented \textsc{Semla}, a novel equivariant message passing architecture exhibiting significantly better efficiency and scalability than existing approaches 3D generation. We have further introduced \textsc{SemlaFlow}, a flow matching model for jointly generating the topology and 3D conformations of molecular graphs. \textsc{SemlaFlow} achieves state-of-the-art results on 3D molecular generation benchmarks with two orders-of-magnitude faster sampling times. We have also highlighted issues with current molecular generation evaluation metrics and proposed the use of energy per atom and strain energy per atom for evaluating the quality of generated molecular conformations.

While we believe our model has made significant progress in solving key challenges for 3D molecular generators, many challenges remain. Firstly, the energies of the molecules generated by \textsc{SemlaFlow} are still significantly higher than that of the dataset; generating molecular coordinates with very high fidelity remains a problem for these models. Including further inductive biases or fine-tuning against an energy model could be an avenue to improve this in future work~\citep{model:boltzmann-generators,dynamics:ito,dynamics:sma-md}. Additionally, while \textsc{SemlaFlow} has shown significant efficiency improvements over existing methods, it still uses a fully-connected message passing component, limiting it's scalability to larger molecular systems. We leave the further enhancement of the scalability of \textsc{Semla} to future work. We believe our model makes crucial step towards the practical application of 3D molecular generators, although we leave the integration of \textsc{SemlaFlow} into drug discovery workflows, either through RL-based fine-tuning or protein pocket conditioned generation, to future work.

\subsubsection*{Acknowledgements}

This work was partially supported by the Wallenberg AI, Autonomous Systems and Software Program (WASP) funded by the Knut and Alice Wallenberg Foundation. Preliminary experiments were enabled by resources provided by the National Academic Infrastructure for Supercomputing in Sweden (NAISS, project: 2024/22-33), partially funded by the Swedish Research Council through grant agreement no. 2022-06725. The authors thank T. Le (Pfizer) for sharing code and weights of~\cite{model:eqgatdiff} ahead of publication.

% \clearpage

\bibliography{references}

\clearpage

\onecolumn

\appendix

\section{ABLATION EXPERIMENTS}
\label{section:ablation-results}

This section includes additional experiments which were designed to explain the design decisions made in the \textsc{Semla} architecture and to assess the benefit of using \textsc{Semla} over other equivariant architectures for molecular generation.

\subsection{Latent Message Size}
\label{section:message-ablation}

Firstly, Table~\ref{table:message-ablation-results} assesses the impact of the \(d_l\) hyperparameter which specifies the size of the latent dimension when performing latent attention. Each model was trained on GEOM Drugs for 100 epochs using the same training setup as the \textsc{SemlaFlow} model in Table~\ref{table:geom-results}.

While \(d_l = 256\) produces the best overall performance in terms of molecule stability, validity and energy, the model takes significantly longer to sample the same number of molecules. Crucially, the difference in performance between all four models is relatively small, despite the notable differences in sample time. For this reason we selected \(d_l = 64\) for all \textsc{SemlaFlow} models.

\begin{table*}[h!]
  \caption{Comparison of \textsc{SemlaFlow} models with different sizes of latent attention dimension \(d_l\). All models are trained in identical conditions for 100 epochs. Sample time is measured by the average number of seconds to generate 1000 molecules. All metrics are averaged over 3 runs.}
  \label{table:message-ablation-results}
  \centering
  \begin{tabular}{lcccccc}
    \toprule
    \(d_l\) & Mol Stab $\uparrow$  & Valid $\uparrow$     & Energy $\downarrow$  
        & Strain $\downarrow$  & Sample Time $\downarrow$   \\
    \midrule
    32      & 97.7$_{\pm \text{0.13}}$  & 92.7$_{\pm \text{0.30}}$
        & 2.73$_{\pm \text{0.014}}$   & 1.80$_{\pm \text{0.004}}$   & 89.0   \\
    64      & 97.6$_{\pm \text{0.02}}$  & 93.1$_{\pm \text{0.30}}$  
        & 2.63$_{\pm \text{0.019}}$   & 1.70$_{\pm \text{0.013}}$   & 98.3   \\
    128     & 97.1$_{\pm \text{0.05}}$  & 92.8$_{\pm \text{0.07}}$
        & 2.87$_{\pm \text{0.026}}$   & 1.98$_{\pm \text{0.022}}$   & 122.0  \\
    256     & 98.4$_{\pm \text{0.06}}$  & 94.6$_{\pm \text{0.14}}$
        & 2.61$_{\pm \text{0.006}}$   & 1.71$_{\pm \text{0.008}}$   & 217.7  \\
    \bottomrule
  \end{tabular}
\end{table*}

\subsection{Architecture Ablation}
\label{section:arch-ablation}

We also wish to compare the generative performance and sampling efficiency of our proposed \textsc{Semla} architecture with existing E(3)-equivariant architectures for 3D molecular generation. To do this we train various network architectures within a consistent experimental setup -- we apply the same flow matching training and inference procedure as \textsc{SemlaFlow} but swap out \textsc{Semla} for existing equivariant architectures. We benchmark \textsc{Semla} against the 4 layer and 9 layer versions of EGNN~\citep{architecture:egnn} proposed in EDM~\citep{model:edm}, as well as the EQGAT network~\citep{architecture:eqgat} used in the state-of-the-art EQGAT-diff model~\cite{model:eqgatdiff}. We also provide a comparison with an expanded, 16 layer version of EGNN, which has roughly the same number of parameters as \textsc{Semla}. In order to handle bond types, we modify the pairwise MLPs for EGNN on the first and last layer of the network in the same way as \textsc{Semla}. Each architecture is given a fixed training budget of 24 hours. Since existing models are not setup for self-conditioned inputs, all models, including \textsc{Semla}, are trained without self conditioning.

\begin{table*}[h!]
  \caption{Comparison of different architectures all trained on an identical flow matching setup. Sample time is measured by the average number of seconds to generate 1000 molecules. All metrics are averaged over 3 runs.}
  \label{table:arch-ablation-results}
  \centering
  \begin{tabular}{lcccccc}
    \toprule
    Architecture   & Mol Stab $\uparrow$  & Valid $\uparrow$     & Energy $\downarrow$  
        & Strain $\downarrow$  & Sample Time $\downarrow$  \\
    \midrule
    EGNN (4 layer)   & 71.4$_{\pm \text{0.76}}$  & 55.1$_{\pm \text{0.06}}$
        & 10.89 $_{\pm \text{0.059}}$  & 10.03$_{\pm \text{0.063}}$  & 69.3  \\
    EGNN (9 layer)   & 94.0$_{\pm \text{0.16}}$  & 87.9$_{\pm \text{0.40}}$
        & 4.04$_{\pm \text{0.047}}$    & 3.13$_{\pm \text{0.023}}$   & 151.4 \\
    EGNN (16 layer)  & 94.7$_{\pm \text{0.06}}$  & 89.9$_{\pm \text{0.06}}$  
        & 3.97$_{\pm \text{0.079}}$    & 3.07$_{\pm \text{0.065}}$   & 532.0 \\
    EQGAT          & 97.1$_{\pm \text{0.07}}$  & 83.9$_{\pm \text{0.33}}$
        & 4.19$_{\pm \text{0.013}}$    & 3.32$_{\pm \text{0.019}}$   & 337.4 \\
    \textsc{Semla} (Ours)          & 96.3$_{\pm \text{0.14}}$  & 91.2$_{\pm \text{0.26}}$
        & 3.04$_{\pm \text{0.014}}$    & 2.15$_{\pm \text{0.014}}$   & 99.4  \\
    \bottomrule
  \end{tabular}
\end{table*}

The results of this architecture ablation study are shown in Table~\ref{table:arch-ablation-results}. \textsc{Semla} shows better validities and significantly better energies than existing architectures, although has slightly lower molecule stability than EQGAT. Crucially though \textsc{Semla} is more than 3 times faster than EQGAT and more than 5 times faster than the similarly-sized 16 layer EGNN network while still producing molecules of comparable or higher quality.

\section{ADDITIONAL RESULTS ON GEOM DRUGS}
\label{section:additional-results}

In this section we include additional results of existing models on GEOM Drugs which we were not able to fit into the main text. We also include the results from Table~\ref{table:geom-results} for full comparison. Many models have been proposed for this task recently and we chose to focus our comparison in the main text to models which learn to generate bonds rather than inferring them from coordinates.

Table~\ref{table:additional-geom-results} shows that while some models which infer bonds (shown in the top segment of the table) produce higher validities than those which do not, their atom and molecule stabilities are often significantly lower. However, since validity only measures whether a molecule can be sanitised by RDKit, we consider molecule stability to be a much better measure of the quality of the generated molecules. We include a further discussion and explanation for this in Appendix~\ref{section:eval-issues}.

\begin{table*}[h!]
  \caption{Additional results on GEOM Drugs, including models which infer bonds based on the generated coordinates. Since many of these models do not publish molecule stability results, numbers marked with * are estimates computed by taking \(AS^{44}\) where \(AS\) is atom stability and 44 is the average number of atoms in GEOM Drugs molecules. This follows the same procedure used in the EquiFM paper to estimate molecule stability.}
  \label{table:additional-geom-results}
  \centering
  \begin{tabular}{lcccccc}
    \toprule
    Model            & Atom Stab $\uparrow$  & Mol Stab $\uparrow$ & Valid $\uparrow$  
        & Unique $\uparrow$  & Novel $\uparrow$  & NFE     \\
    \midrule
    EDM              & 81.3  & 0.0*  & --    & --  & --  & 1000      \\
    GCDM             & 89.0  & 5.2   & --    & --  & --  & 1000      \\
    MUDiff           & 84.0  & 60.9  & 98.9  & --  & --  & 1000      \\
    GFMDiff          & 86.5  & 3.9   & --    & --  & --  & 500       \\
    EquiFM           & 84.1  & 0.0*  & 98.9  & --  & --  & --        \\
    GeoBFN           & 86.2  & 0.0*  & 91.7  & --  & --  & 2000        \\
    GeoLDM           & 98.9  & 61.5* & \textbf{99.3}  & --  & --  & 1000        \\
    \midrule
    FlowMol          & 99.0           & 67.5  & 51.2  & --               & --             & 100      \\
    MiDi             & \textbf{99.8}  & 91.6  & 77.8  & \textbf{100.0}  & \textbf{100.0}  & 500      \\
    EQGAT-diff       & \textbf{99.8}$_{\pm \text{0.0}}$  & 93.4$_{\pm \text{0.21}}$  & 94.6$_{\pm \text{0.24}}$
        & \textbf{100.0}$_{\pm \text{0.0}}$  & 99.9$_{\pm \text{0.07}}$  & 500      \\
    \textsc{SemlaFlow} (Ours) & \textbf{99.8}$_{\pm \text{0.0}}$  & \textbf{97.3}$_{\pm \text{0.08}}$  & 93.9$_{\pm \text{0.19}}$
        & \textbf{100.0}$_{\pm \text{0.0}}$  & 99.6$_{\pm \text{0.03}}$  & 100      \\
    \bottomrule
  \end{tabular}
\end{table*}

\section{FURTHER MODEL AND TRAINING DETAILS}
\label{section:training-appendix}

This section provides further detail on the design of the full \textsc{Semla} architecture, as well further training details including the hyperparameters used to trian QM9 and GEOM Drugs models.

\subsection{The \textsc{Semla} Architecture}

\begin{figure}
    \centering
    \includegraphics[width=0.22\linewidth]{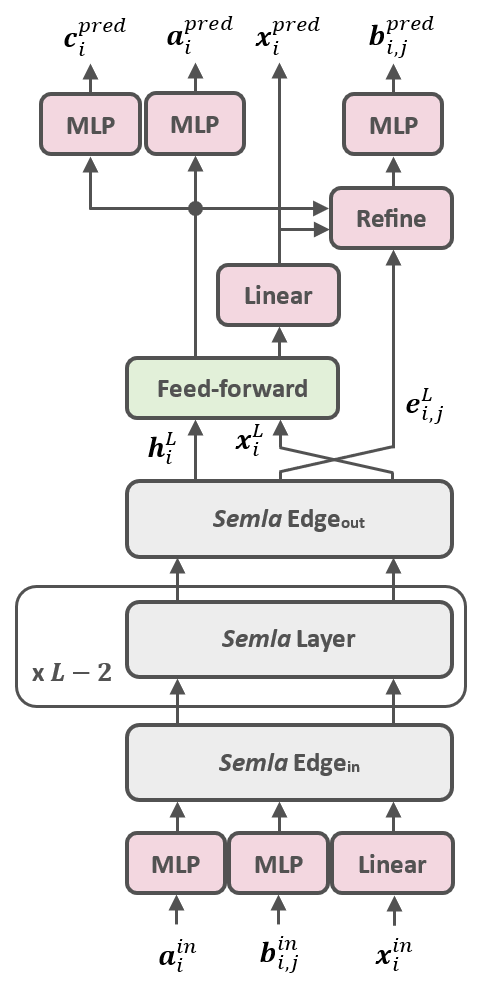}
     \caption{An overview of a full \textsc{Semla} model. A \textsc{Semla} model is created from a stack of \textsc{Semla} layers where the first layer encodes bond information into the node features and the final layer generates bond information from the pairwise features.}
     \label{fig:model-overview}
\end{figure}

As mentioned in Section~\ref{section:semla}, we do not carry pairwise edge features through the model, but rather encode the edge information into the node features on the first layer and then generate edge features on the final layer. After these initial edge features are generated they are passed through a learnable refinement component where the final predicted coordinates are also given as input, along with the final invariant node features. The remaining components of the model are used to encode atom and bond types and predict distributions for atom types, bond types and formal charges. A full overview of a \textsc{Semla} model is shown in Figure~\ref{fig:model-overview}.

\subsection{Training Details}

All models were trained with the AMSGrad~\citep{dl:amsgrad} variant of the Adam optimiser~\citep{dl:adam} with a learning rate (LR) of 0.0003. We also apply linear LR warm-up, using 2000 warm-up steps for QM9 and 10000 warm-up steps for GEOM Drugs. During training we clip the norms of the gradients at 1.0 for all models. Loss weightings \((\lambda_x, \lambda_a, \lambda_b, \lambda_c) = (1.0, 0.2, 0.5, 1.0)\) were used to train QM9 models. The same weightings were used for GEOM Drugs, except we set \(\lambda_b = 1.0\).

When training with self-conditioning half of the training batches are treated as normal and the other half are trained on as self-conditioning batches. In this case the batch is firstly processed by the model to generate conditioning inputs, these are then detached from the computation graph, and finally used as conditioning inputs for the model training step. In practice the conditioning inputs are concatenated with the interpolated data and embedded at the start of the model. For atom and bond types the conditioning inputs are softmax-normalised probability distributions over the predicted categorical types.

In order to make the training as efficient as possible we place molecules in the dataset into buckets based on their size, and then form minibatches for training and evaluation within the buckets. This ensures that all batches contain similarly sized molecules so that the amount of padding within each batch is minimised. With this setup we can also apply a cost function to select the batch size for each bucket separately; since the amount of memory required to process a molecule increases quadratically with the number of atoms, this helps to balance the GPU memory consumption for each batch. In practice, though, we simply apply a linear cost function and use a batch size of 4096 atoms per batch for all \textsc{SemlaFlow} models. While we have found our bucketing scheme leads to a significant increase in training speed, it may also introduce additional bias into the training since molecules within each batch are no longer selected completely at random. Although we have not attempted to quantify this bias our results seem to show that bucketing is not significantly detrimental to performance.

\section{EVALUATION METRICS}
\label{section:eval-issues}

In this section we outline a number of shortcomings with current evaluation for metrics for unconditional 3D generative models. Firstly, we provide a full description for each existing benchmark metric we have used:
\begin{itemize}
    \item \textbf{Atom stability} measures the proportion of atoms which have the correct number of bonds, according to a pre-defined lookup table.
    \item \textbf{Molecule stability} then measures the proportion of generated molecules for which all atoms are stable.
    \item \textbf{Validity} is given by the proportion of molecules which can be successfully sanitised using RDKit.
    \item \textbf{Uniqueness} measures the proportion of generated molecules which are unique. This is calculated by comparing molecules based on their canonical SMILES representation.
    \item \textbf{Novelty} measures the proportion of generated molecules which are not in the training set.
\end{itemize}

\subsection{Issues with Existing Metrics}

While a lot of progress has been made recently in improving benchmarks for structure-based drug design~\citep{sbdd:benchmarking,sbdd:posebusters}, we believe many evaluation metrics used for unconditional 3D generation are still not fit for purpose. We hope that the introduction of energy per atom and strain energy per atom will help to alleviate this but believe it is still important to discuss the shortcomings of existing evaluation methods. In this section we highlight issues with the standard benchmark metrics for this task.

\paragraph{Validity}

\begin{figure}
    \centering
    \includegraphics[width=0.9\linewidth]{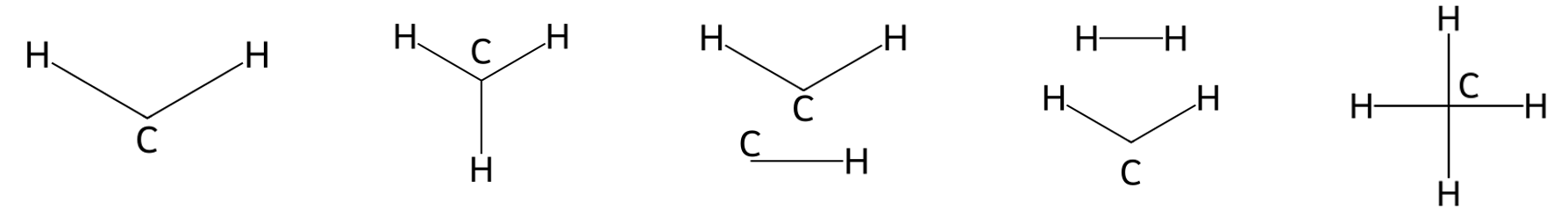}
     \caption{Examples of molecules which would be considered valid by RDKit.}
     \label{fig:valid-mols}
\end{figure}

Since RDKit is free to add implicit hydrogens and, in some cases, to modify the formal charge on atoms, RDKit validity tells us very little about the "correctness" of a molecule. This is shown explicitly in Figure~\ref{fig:valid-mols}; all molecules shown in the figure would be marked as valid by RDKit, even though only the rightmost molecule has the correct number of bonds. Additionally, the validity metric on its own is not capable of flagging disconnected molecules -- generated molecules which contain multiple fragments which are not connected by any bond. The validity metric is, however, capable of spotting when a molecule has too many bonds.

\paragraph{Atom and Molecule Stability}

Different models sometimes use different lookup tables to compute atom (and therefore molecule) stability, potentially leading to different evaluation results for the same generated molecules. Often this depends on whether models predict formal charges for each atom or not, since those that do need to take the charge into account in the look-up table. Additionally, we have found that existing lookup tables used to define atom stability have little basis in chemical validity. As an example, the existing definition for atom stability allows an uncharged carbon atom with 3 (single covalent bond equivalent) bonds to be considered 'stable', although this has little-to-no chemically valid basis.

\paragraph{Conclusion}

We believe the issues highlighted here reflect the importance of using metrics such as energy which is able to quantify the quality of both the generated topology as well as the generated molecular conformation. We also hope that highlighting these issues will lead to more in-depth evaluations of 3D unconditional molecular generation models in the future and lay a foundation for more reproducible and coherent benchmarking.

\newpage

\section{SAMPLES FROM SEMLAFLOW}
\label{section:samples}

In this section we present samples from \textsc{SemlaFlow} trained on GEOM Drugs. The samples were generated randomly but we have rotated them where necessary to aid visualisation.

\begin{figure}[ht!]
    \centering
    \begin{subfigure}[t]{0.22\linewidth}
        \centering
        \includegraphics[width=\textwidth]{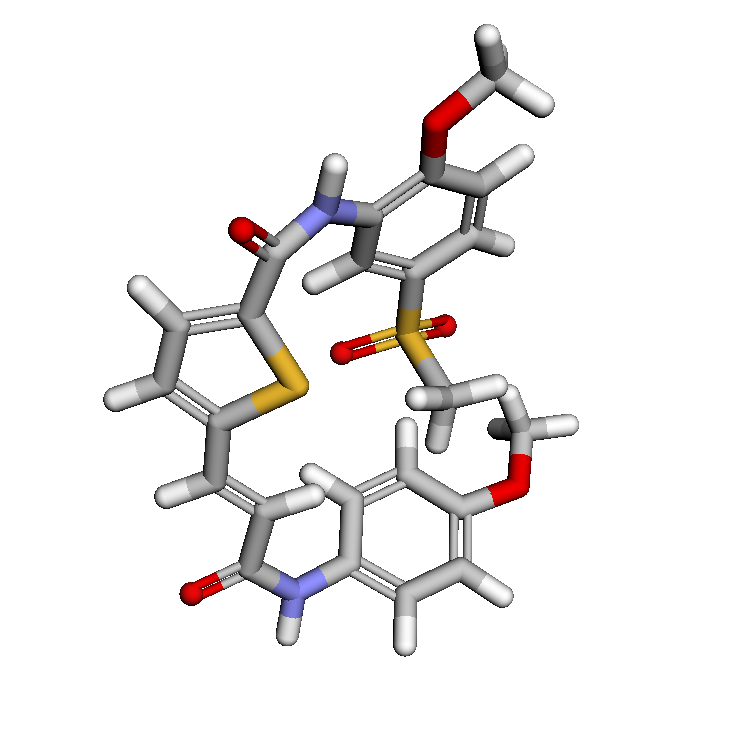}
    \end{subfigure}
    \begin{subfigure}[t]{0.22\linewidth}
        \centering
        \includegraphics[width=\textwidth]{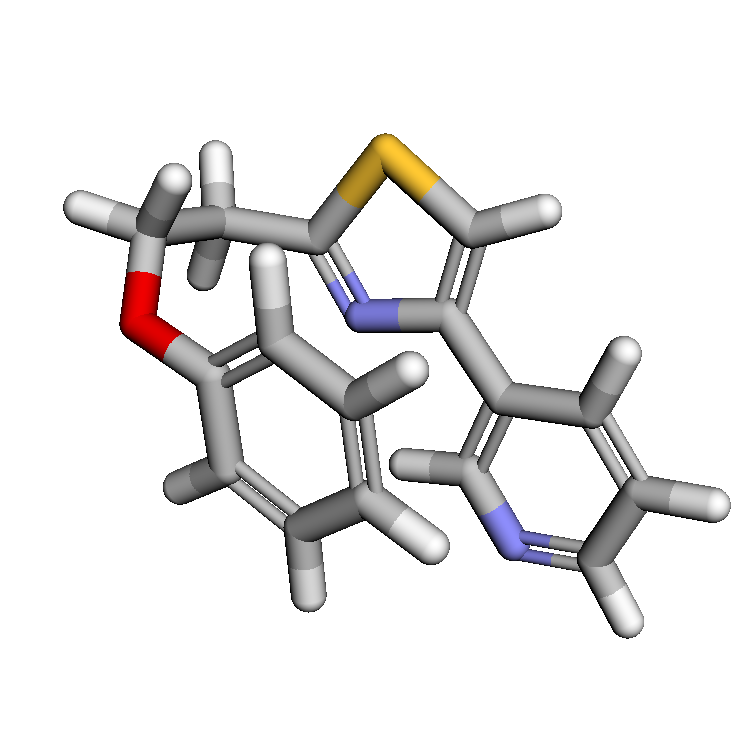}
    \end{subfigure}
    \begin{subfigure}[t]{0.22\linewidth}
        \centering
        \includegraphics[width=\textwidth]{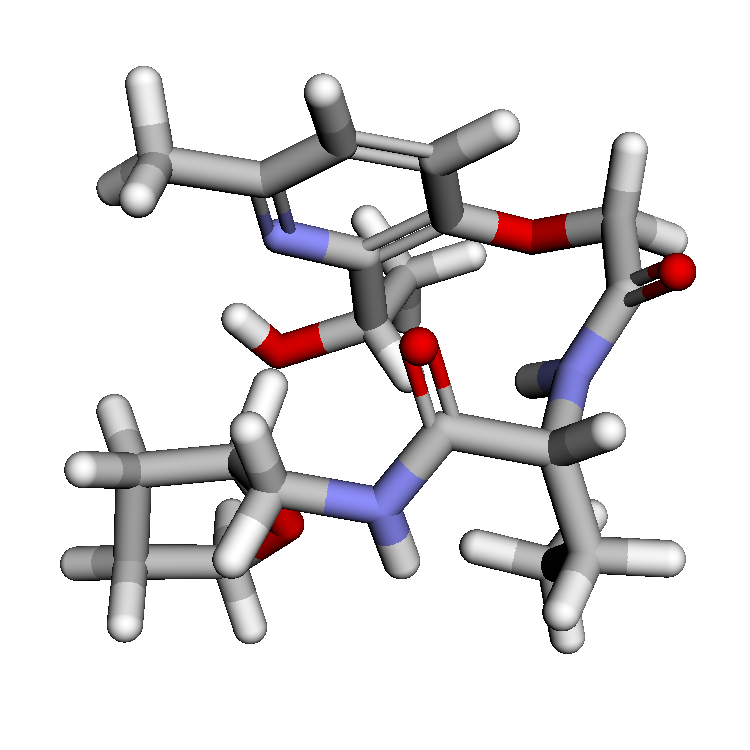}
    \end{subfigure}
    \begin{subfigure}[t]{0.22\linewidth}
        \centering
        \includegraphics[width=\textwidth]{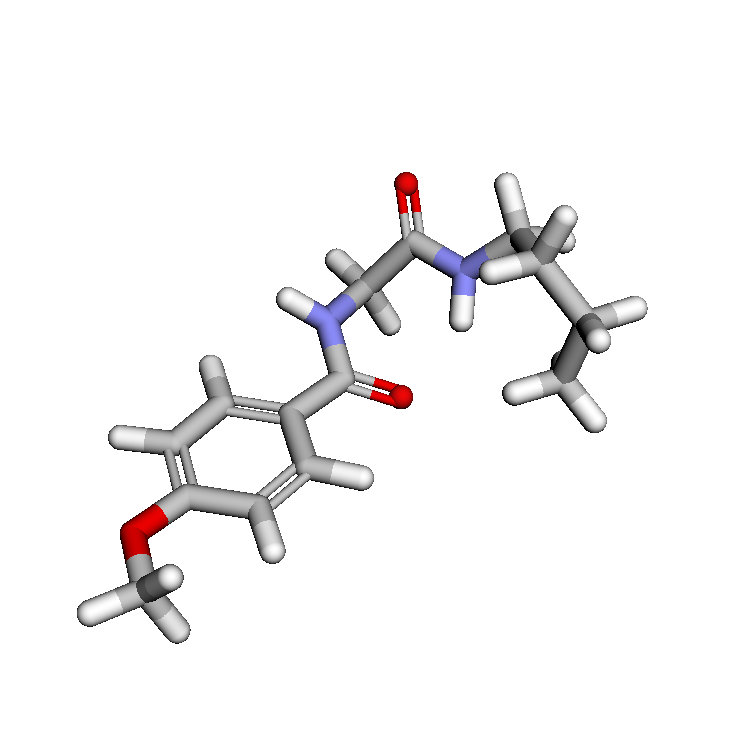}
    \end{subfigure}
    \begin{subfigure}[t]{0.22\linewidth}
        \centering
        \includegraphics[width=\textwidth]{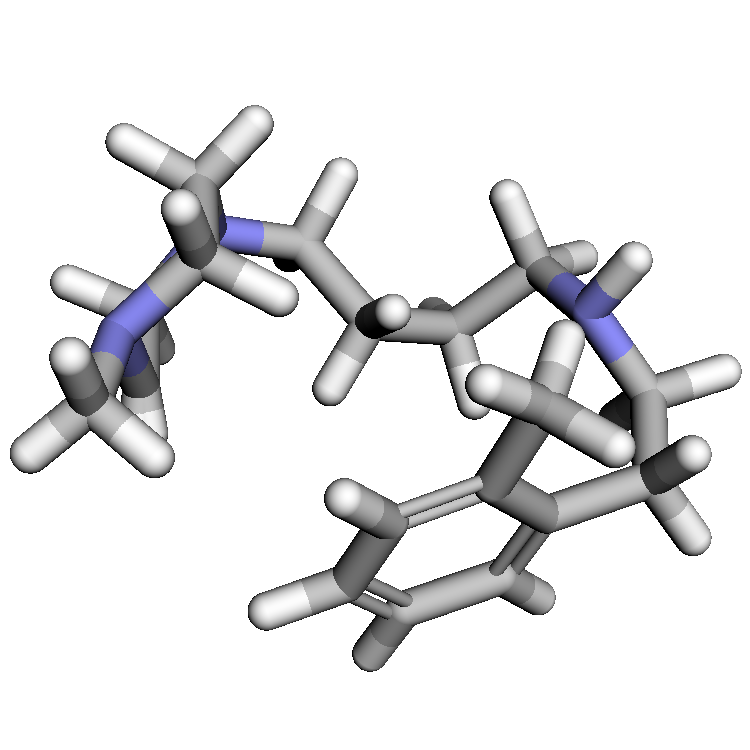}
    \end{subfigure}
    \begin{subfigure}[t]{0.22\linewidth}
        \centering
        \includegraphics[width=\textwidth]{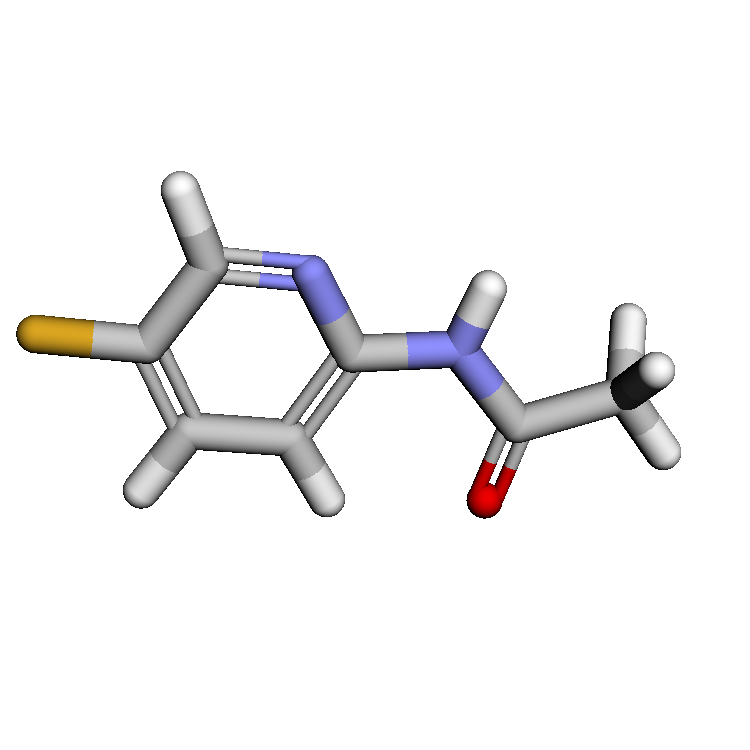}
    \end{subfigure}
    \begin{subfigure}[t]{0.22\linewidth}
        \centering
        \includegraphics[width=\textwidth]{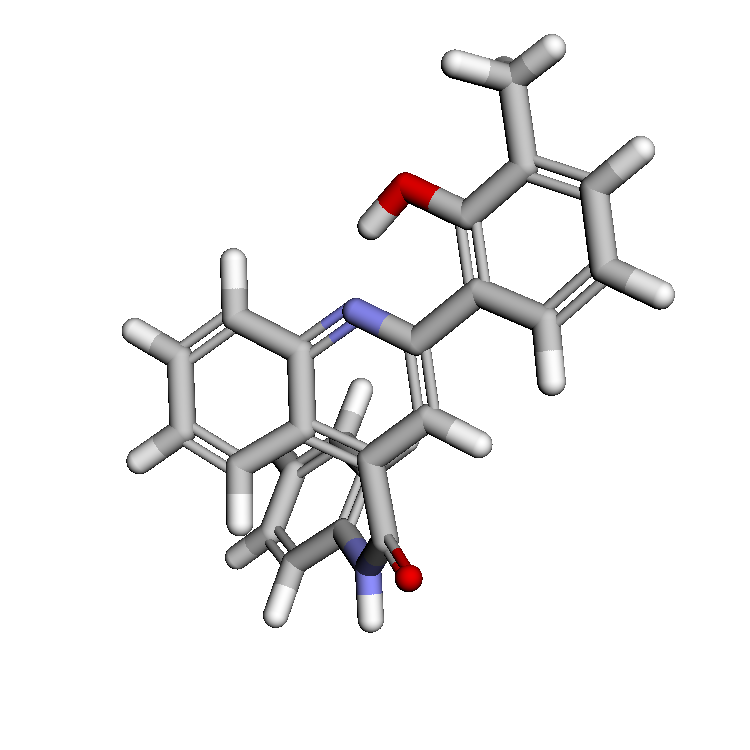}
    \end{subfigure}
    \begin{subfigure}[t]{0.22\linewidth}
        \centering
        \includegraphics[width=\textwidth]{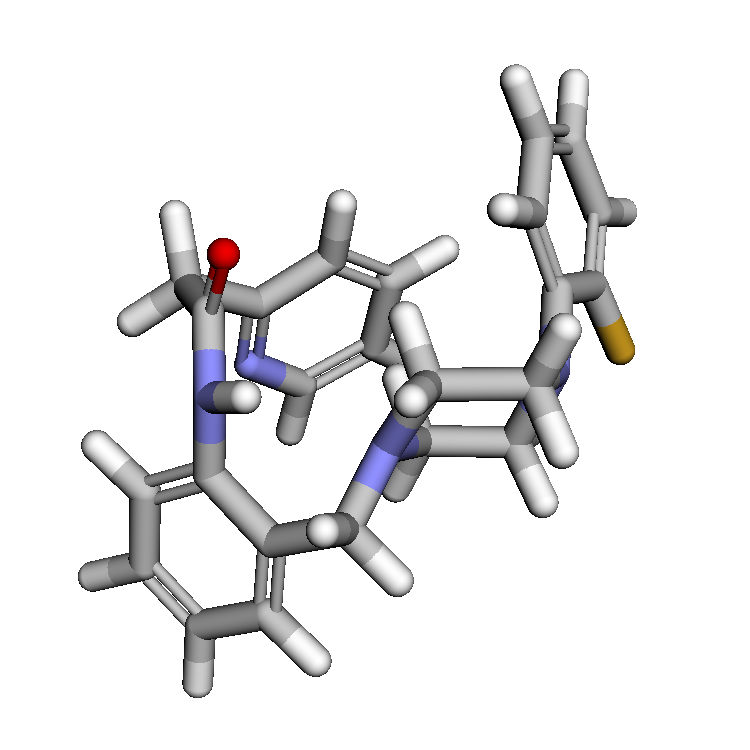}
    \end{subfigure}
    \begin{subfigure}[t]{0.22\linewidth}
        \centering
        \includegraphics[width=\textwidth]{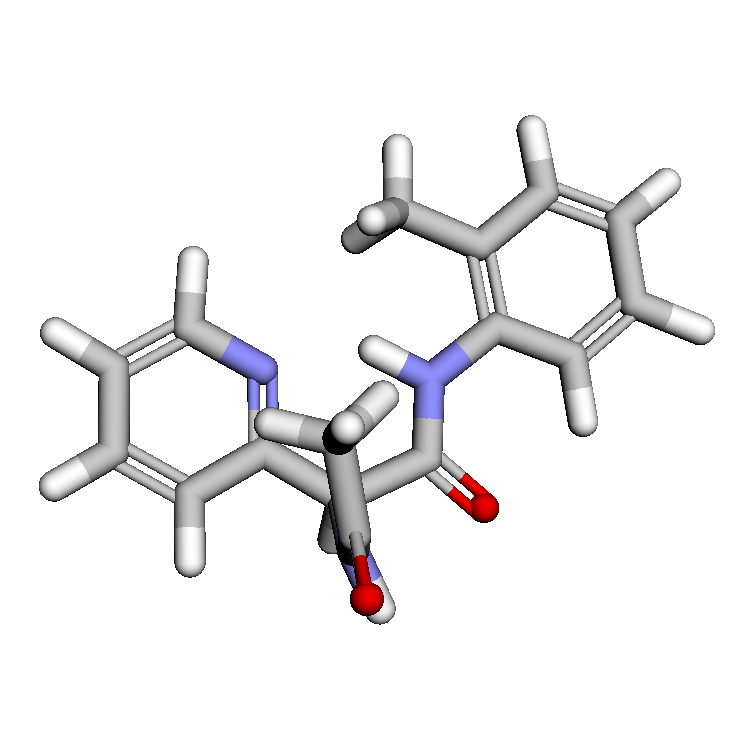}
    \end{subfigure}
    \begin{subfigure}[t]{0.22\linewidth}
        \centering
        \includegraphics[width=\textwidth]{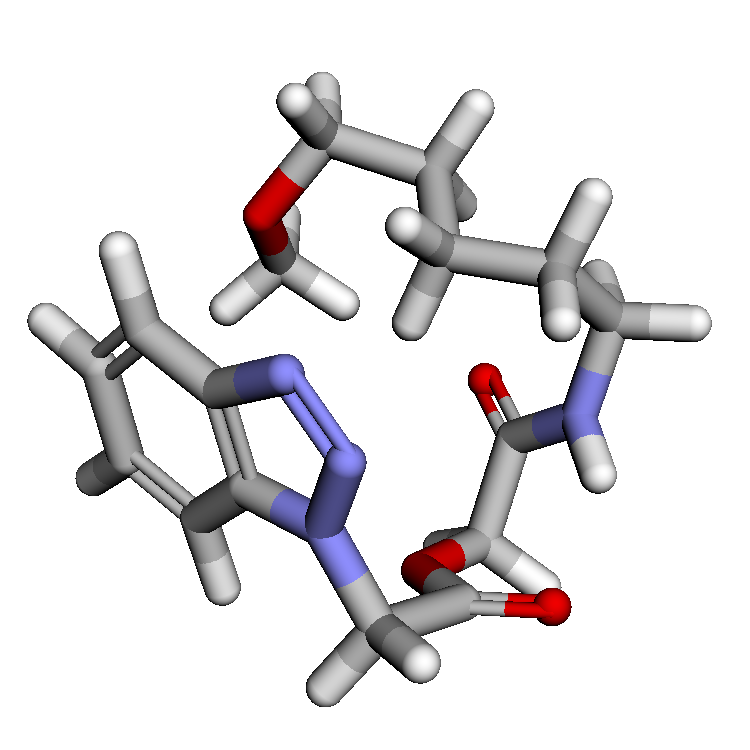}
    \end{subfigure}
    \begin{subfigure}[t]{0.22\linewidth}
        \centering
        \includegraphics[width=\textwidth]{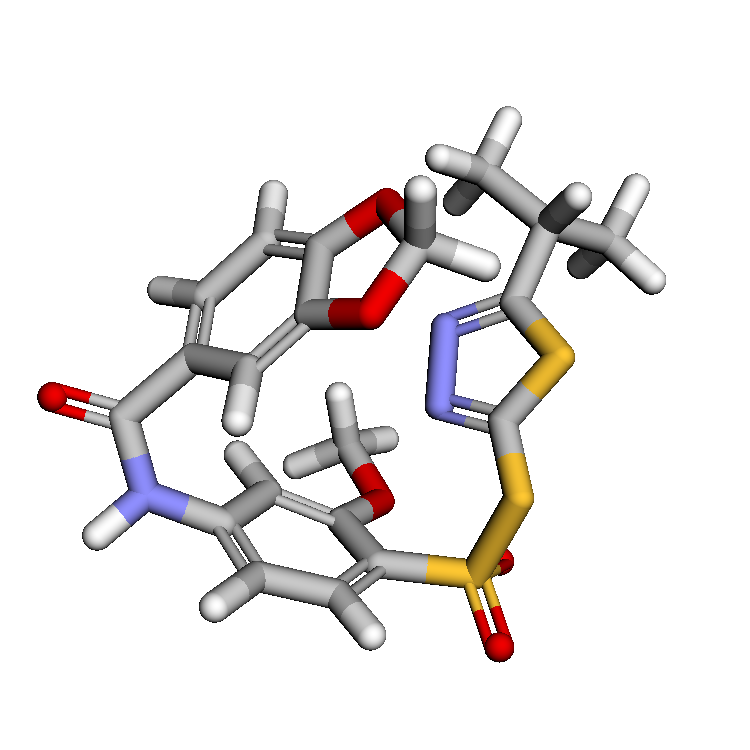}
    \end{subfigure}
    \begin{subfigure}[t]{0.22\linewidth}
        \centering
        \includegraphics[width=\textwidth]{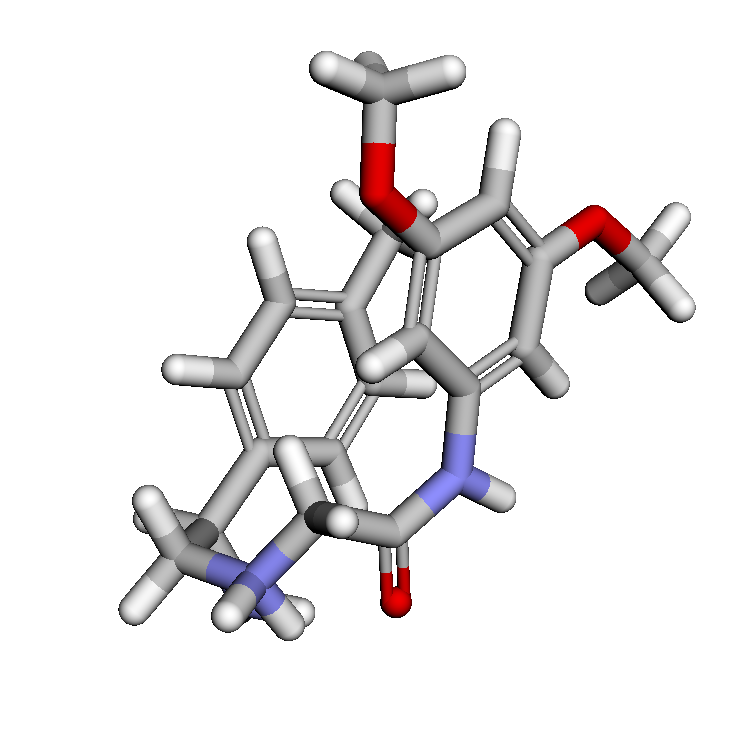}
    \end{subfigure}
    \begin{subfigure}[t]{0.22\linewidth}
        \centering
        \includegraphics[width=\textwidth]{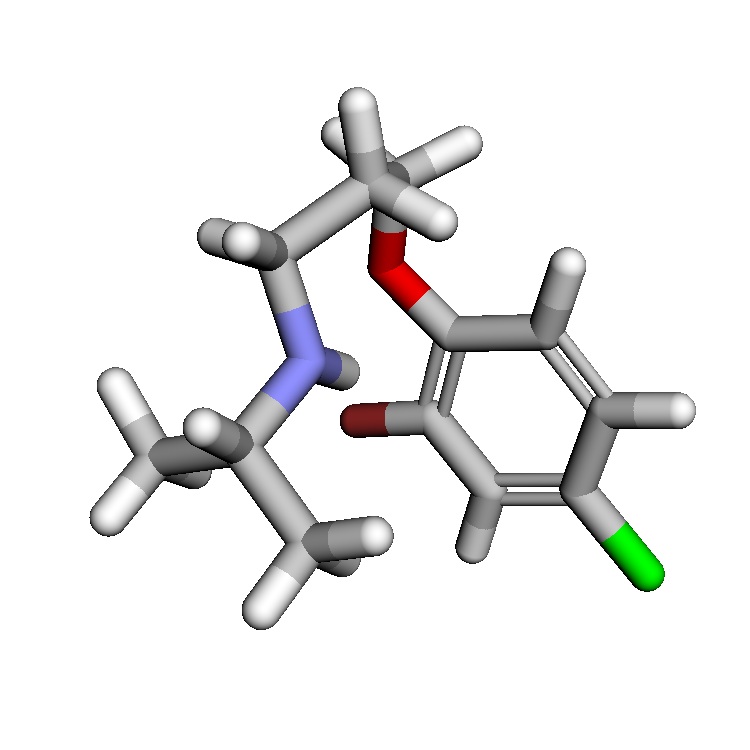}
    \end{subfigure}
    \begin{subfigure}[t]{0.22\linewidth}
        \centering
        \includegraphics[width=\textwidth]{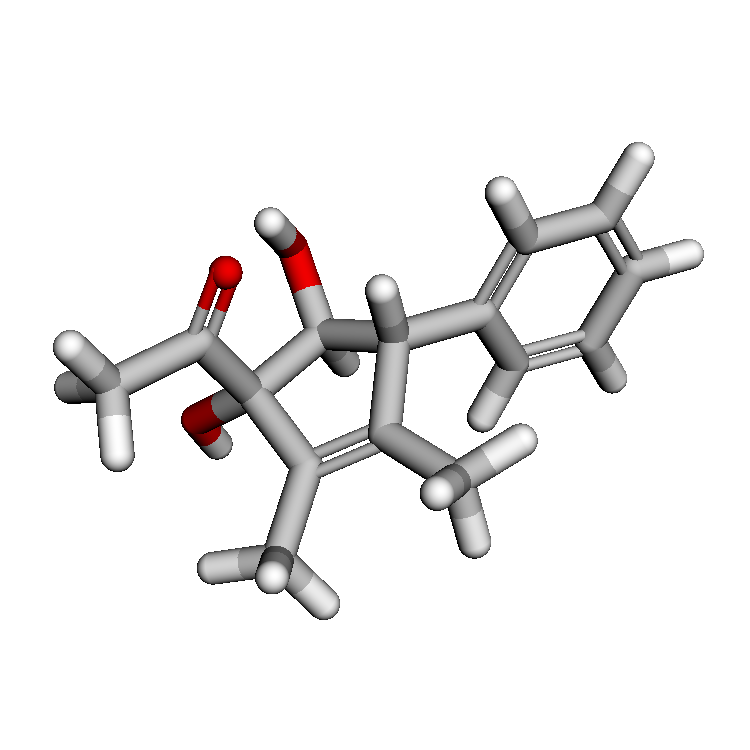}
    \end{subfigure}
    \begin{subfigure}[t]{0.22\linewidth}
        \centering
        \includegraphics[width=\textwidth]{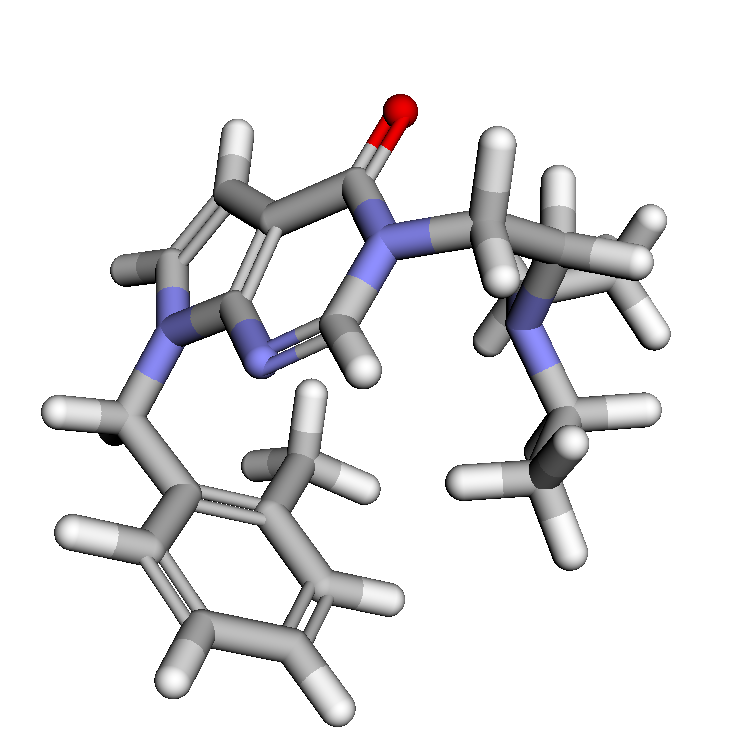}
    \end{subfigure}
    \caption{Random samples from a \textsc{SemlaFlow} model trained on GEOM Drugs. These samples were generated using 100 ODE integration steps.}
    \label{fig:model-samples}
\end{figure}

%%%%%%%%%%%%%%%%%%%%%%%%%%%%%%%%%%%%%%%%%%%%%%%%%%%%%%%%%%%%

\end{document}